\documentclass[letterpaper, 10pt, conference]{ieeeconf}  

\IEEEoverridecommandlockouts  
\usepackage[utf8]{inputenc}   
\usepackage{amsmath, amssymb} 
\usepackage{booktabs}         
\usepackage{graphicx}         

\title{\LARGE \bf
    Frequency-Modulated Piezoelectric Haptic Display
}

\author{Boyuan Liang, Lingfeng Sun, Masayoshi Tomizuka
\thanks{*This work was supported by FANUC Advanced Research Laboratories.}%
\thanks{All authors are with the Department of Mechanical Engineering, University of California, Berkeley, CA, USA {\tt\small \{liangb, lingfengsun, tomizuka\}@berkeley.edu}}%
}

\begin{document}

\maketitle
\thispagestyle{empty}
\pagestyle{empty}

\begin{abstract}
We present a frequency-modulated (FM) haptic display based on piezoelectric vibrating actuators. Existing haptic displays commonly encode haptic intensity through the deformation amplitude of individual haptic pixels. Although amplitude-modulated (AM) approaches have enabled compact haptic pixels, independently controlling the deformation amplitude of a large number of pixels can require increasingly complex and bulky driving systems, posing challenges for scaling toward high-density, large-area wearable displays. To address this scaling challenge, we investigate an FM design principle in which haptic intensity is encoded through vibration frequency. We further develop a \textit{Shared-Source Frequency Modulation} (SSFM) structure in which multiple haptic pixels are powered by a common power amplifier while their vibration spectrum are controlled individually, reducing the need for independent high-power amplification at each pixel. A proof-of-concept piezoelectric haptic display was built and evaluated on rendering spatial and temporal haptic patterns through volunteer tests. The results show that participants reliably distinguished spatial and temporal patterns encoded using FM principles within the investigated operating range. These findings demonstrate the feasibility of FM-based distributed haptic rendering and suggest a potential pathway toward more compact driving architectures for future high-density, large-area wearable haptic displays.
\end{abstract}

\section{Introduction}
\label{sec:intro}

Haptic displays enable digital information to be conveyed through spatially and temporally varying tactile stimulation. Analogous to pixels in a visual display, these systems often employ arrays of individually controllable surface elements, or haptic pixels, whose mechanical responses can be modulated to render tactile patterns over a contact surface. Because the displayed information can be dynamically programmed, haptic displays have been explored as general-purpose tactile interfaces for applications including refreshable Braille and tactile graphics \cite{watanabe2006practical, russomanno2015refreshing}, interactive tactile interfaces \cite{matsushita1997holowall, guimbretiere2001fluid}, and virtual and augmented reality (VR/AR) \cite{benko2016normaltouch, wang2019multimodal}.

Considerable progress has been made in miniaturizing haptic pixels and improving their response speed. As haptic displays move toward higher pixel densities and larger interactive areas, however, scaling the driving architecture becomes increasingly important in addition to scaling the actuators themselves. In many existing systems, independently controlling the mechanical amplitude of each pixel requires dedicated amplification, routing, or energy-delivery resources. Consequently, a display composed of small actuators does not necessarily result in a compact distributed system. As the pixel count increases, the electronics and interconnections required near the actuation surface can become a significant part of the overall device.

This challenge is particularly relevant for wearable haptic displays, where the components located near the body must remain compact, lightweight, and mechanically unobtrusive. In this paper, we distinguish between \textit{on-skin} and \textit{off-skin} components of the driving system. We define the on-skin driver as the circuitry and interconnections that must remain physically close to the haptic pixels, whereas centralized power supplies, amplifiers, or other remote energy sources are considered off-skin components. From this perspective, an important system-level question is whether independently controllable haptic pixels can be realized while minimizing the amount of high-power driving hardware that must scale with the number of pixels.

We investigate a frequency-modulated (FM) approach to distributed haptic rendering that is intended to reduce this on-skin driving requirement. The approach is motivated by the well-established frequency dependence of human vibrotactile sensitivity. Psychophysical studies have shown that tactile displacement thresholds vary substantially with vibration frequency and can reach sub-micrometer levels near the region of highest sensitivity around $200,\mathrm{Hz}$ \cite{brisben1999detection}. This frequency dependence suggests an alternative to controlling perceived vibration strength primarily through mechanical amplitude. Within an appropriate operating range, frequency can also serve as a control variable for producing distinguishable levels of tactile intensity while maintaining relatively small displacement amplitudes.

Building on this observation, we propose a Shared-Source Frequency Modulation (SSFM) architecture. Multiple piezoelectric haptic pixels share a common high-voltage power amplifier, while compact multichannel analog switches independently control the effective vibration frequency delivered to each pixel. The high-power amplification stage can therefore remain off-skin, while the pixel-dependent on-skin driver consists primarily of switching circuitry and electrical interconnections. Because multiple high-voltage switching channels can be integrated within compact electronic packages, this architecture provides a potential route toward increasing pixel count without requiring a corresponding number of local power amplifiers.

The objective of this work is to determine whether such a shared-amplifier FM architecture can reduce pixel-level driving requirements while preserving the ability to communicate useful spatial, temporal, and intensity information through touch. We develop a proof-of-concept prototype and conduct human-participant experiments evaluating recognition of spatial-temporal haptic patterns and discrimination of FM-encoded intensity levels. The results show that participants can reliably distinguish the tested patterns and intensity levels, providing experimental evidence for the feasibility of FM-based distributed haptic rendering.

The contributions of this work are therefore threefold. First, we introduce frequency modulation as a system-level design strategy for distributed haptic displays, motivated by the frequency dependence of human vibrotactile sensitivity. Second, we develop the SSFM architecture, which enables multiple piezoelectric pixels to share a high-voltage amplification source while retaining independent pixel-level frequency control. Third, we experimentally evaluate whether the resulting system can convey distinguishable spatial-temporal patterns and intensity levels to human users. The present study focuses on validating the modulation principle and shared-source architecture rather than realizing a complete high-density wearable device. We discuss a possible pathway toward such an implementation in Section \ref{sec:discussion}, which remains an important direction for future work.

\section{Related Works}
\label{sec:related}

\subsection{High-Resolution Haptic Displays}

High-fidelity haptic displays require both sufficient spatial resolution and sufficiently rapid temporal response. Perceptual studies suggest that refresh intervals below approximately $50$ ms can help maintain temporal continuity \cite{craig1987vibrotactile}, while spatial resolutions on the order of a few millimeters can be important for resolving fine tactile patterns \cite{copeland2010identification, kern2023engineering}. Substantial progress toward these requirements has been achieved using a diverse range of actuation technologies.

Millimeter-scale haptic pixels have been demonstrated using mechanical \cite{sarakoglou2012high}, pneumatic \cite{heisser2025explosion}, thermal \cite{besse2017flexible}, and electromagnetic \cite{zarate2017keep, kim2020braille} mechanisms. Rapid actuation has similarly been achieved using electrostatic \cite{grasso2023fully}, piezoelectric \cite{jin2022highly}, and combustion-based \cite{heisser2021valveless} approaches. Collectively, these studies have substantially advanced the achievable spatial resolution, response speed and force output of tactile displays.

Many such systems modulate tactile output through the displacement or vibration amplitude of individual pixels. This amplitude-modulated (AM) principle provides direct and effective control over the mechanical output of each actuator and has enabled a wide range of successful haptic interfaces. At the same time, when AM architectures are extended toward large numbers of independently controlled pixels, the associated driving hardware becomes an additional design consideration. Depending on the actuation mechanism, individual pixels may require dedicated amplification channels, valves, transmission paths, or other routing resources. Thus, further miniaturization of the actuators alone does not necessarily reduce all components that must be distributed across a large-area haptic interface.

\subsection{Shared Actuation Resources for Large-Scale Displays}

Recent work has begun to explore architectures in which actuation resources are shared among multiple haptic pixels, offering promising alternatives to fully replicated pixel-level driving hardware. Linnander et al. \cite{linnander2025tactile}, for example, demonstrated an optically driven tactile display in which a shared laser source is rapidly steered among photoabsorbing elements to produce localized thermal actuation. This approach substantially reduces the need for dedicated high-power actuation hardware at individual pixels and enables arrays containing thousands of tactile elements.

Shen et al. \cite{shen2023fluid} demonstrated a complementary electrohydraulic strategy in which electrostatic forces squeeze liquid to create localized surface deformation. Rather than assigning an independent power stage to each tactile location, the architecture exploits shared electrical resources to generate spatially distributed tactile outputs.

These systems illustrate an important direction in scalable haptic display design. The actuation resource itself can be shared rather than replicated at every pixel. Optical and fluidic approaches are particularly effective when their corresponding energy-transmission mechanisms can be integrated into the target form factor. For densely distributed wearable electronics, however, it is also useful to investigate architectures that retain an entirely electrical interface between the remote power source and the local actuators. Thin flexible conductors and compact integrated switching devices are already widely compatible with wearable electronic fabrication, motivating the exploration of a complementary electrically driven solution.

\begin{figure*}[t]
    \centering
    \includegraphics[width=\linewidth]{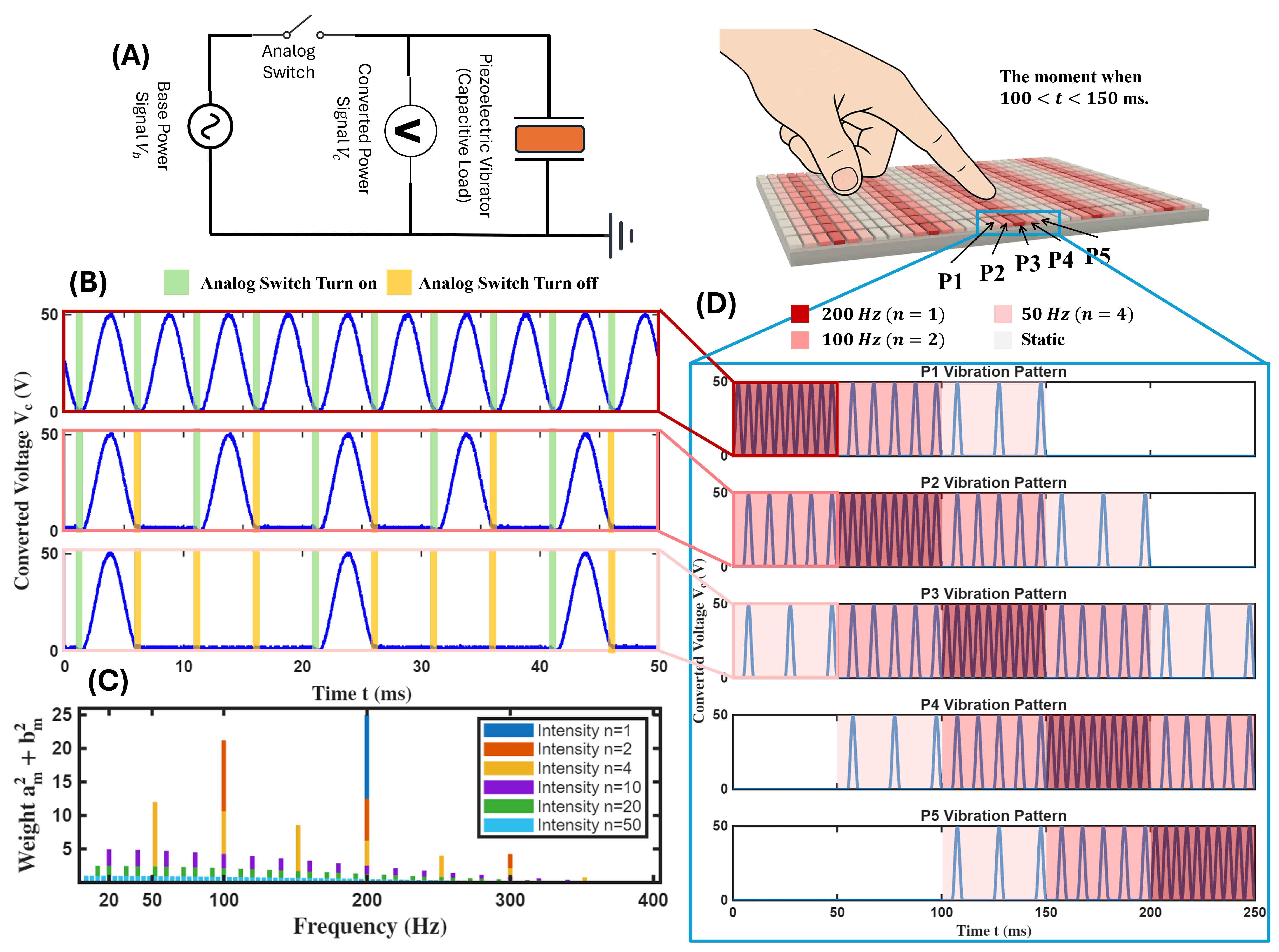}
    \caption{Overview of the SSFM structure, illustrated with an example in which the base frequency is $f_0=200\, Hz$, the peak drive voltage is $V_0=50\, V$, and the refresh interval is $\Delta t=50\, ms$. (A) Schematic of a single pixel, powered by the shared base signal and gated by an analog switch. The switch changes state only while $V_b<\delta$. (B) Temporal relation among the base waveform $V_b$, the analog switch action, and the converted waveform. To produce a nominal frequency of $f_0/n$, the switch is opened for $n-1$ base periods and then closed again for one period. (C) Spectra of the converted waveform for selected pixel intensities $n=1,2,4,10,20,50$. As $n$ increases and the nominal frequency decreases, the dominant component remains at $f_0/n$, while a substantial high-frequency portion is retained. This retained content may help keep low-nominal-frequency vibrations perceptible. (D) Example of how dynamic haptic patterns are generated by SSFM, where darker shading denotes a higher nominal frequency. Assigning a different intensity sequence to each of the pixels P1 to P5 renders a line sweeping from left to right. The sequences are P1: $\{1, 2, 4, \infty, \infty\}$, P2: $\{2, 1, 2, 4, \infty\}$, P3: $\{4, 2, 1, 2, 4\}$, P4: $\{\infty, 4, 2, 1, 2\}$, and P5: $\{\infty, \infty, 4, 2, 1\}$.}
    \label{fig:fm}
\end{figure*}

\section{Shared-Source Frequency Modulation}
\label{sec:fm}

This section describes the SSFM structure, the mechanism by which it realizes independently controlled vibration frequencies at individual pixels, and the way dynamic haptic patterns are composed. We also present a spectral analysis of the resulting output waveform together with a first-order estimate of its thermal behavior.

\subsection{Single-Pixel Frequency Modulation}
\label{subsec:fm-ssfm}
The off-skin driver supplies a single sinusoidal base waveform $V_b$ at a fixed frequency $f_0$, which is shared by all pixels. The on-skin driver converts this common waveform into a set of desired nominal frequencies $f\in\{f_i\}_{i=1}^M$, one per pixel, by gating the shared signal with analog switches. We use the term \textit{nominal} because the resulting output is not a simple harmonic signal but a composite periodic signal of period $1/f$, whose spectral content is examined later in this subsection.

We take the base waveform to be a biased sinusoid spanning $0$ to $V_0$, a common drive pattern for piezoelectric vibrators that avoids negative voltages,
\begin{equation}
    V_b(t)=\frac{V_0}{2}(1-\cos\omega t),\quad \omega=2\pi f_0
\label{eq:base-wave}
\end{equation}

The base waveform $V_b(t)$ is applied to the input side of the analog switch array, as shown in Fig.~\ref{fig:fm}A, together with a low-voltage-triggered controller. Whenever $V_b(t)$ falls below a threshold $\delta$, the controller issues microsecond-scale control signals that set the state of each switch channel according to the desired output. Restricting switching to these low-voltage instants keeps the voltage discontinuity at each transition small, which is important for the thermal behavior discussed at the end of this subsection. To obtain a nominal frequency of $f_0/2$, the controller opens the corresponding switch and closes it again at the following trigger instant; to obtain $f_0/4$, it keeps the switch open for two base periods before closing it. Fig.~\ref{fig:fm}B illustrates the temporal relation among the base waveform, the control signal, and the resulting output. In this way the base waveform can be converted into an output waveform of nominal frequency $f_0/n$ for any positive integer $n$. We denote the converted waveform by $V_c(t)$ and refer to $n$ as the \textit{pixel intensity}; a smaller $n$ corresponds to a higher nominal frequency and, given the frequency dependence of vibrotactile sensitivity, to a stronger perceived vibration. The limiting case $n\to\infty$ gives $V_c(t)\equiv 0$, for which the haptic pixel remains static. Because the switch passes or blocks entire base periods, $V_c(t)$ retains spectral content well above its nominal frequency, including components at and above $f_0$, and is therefore not a simple harmonic waveform. The waveform corresponding to intensity $n$ can be written as
\begin{equation}
    V_{c,n}(t)=\begin{cases}
        \frac{V_0}{2}(1-\cos\omega t), & \quad \frac{kn}{f_0}\leq t < \frac{kn+1}{f_0}\\
        0, & \quad \frac{kn+1}{f_0}\leq t < \frac{kn+n}{f_0}
    \end{cases}
\end{equation}
where $k$ is any nonnegative integer. Expanding $V_{c,n}(t)$ in a Fourier series gives
\begin{equation}
    V_{c,n}(t)=\frac{V_0}{2n}+\sum_{m=1}^\infty\Big[a_m\cos\Big(\frac{\omega mt}{n}\Big)+b_m\sin\Big(\frac{\omega mt}{n}\Big)\Big]
\end{equation}
where the coefficients admit the closed forms
\begin{equation}
    a_m=\frac{V_0n^2\sin\frac{2\pi m}{n}}{2\pi m(n^2-m^2)}, \quad
    b_m=\frac{V_0n^2\sin^2\frac{\pi m}{n}}{\pi m(n^2-m^2)}
\end{equation}

Both coefficients decay asymptotically as $m^{-3}$, so the amplitude of high-order harmonics falls off rapidly and the spectrum remains concentrated at the lower harmonic orders.

\begin{figure*}[t]
    \centering
    \includegraphics[width=\linewidth]{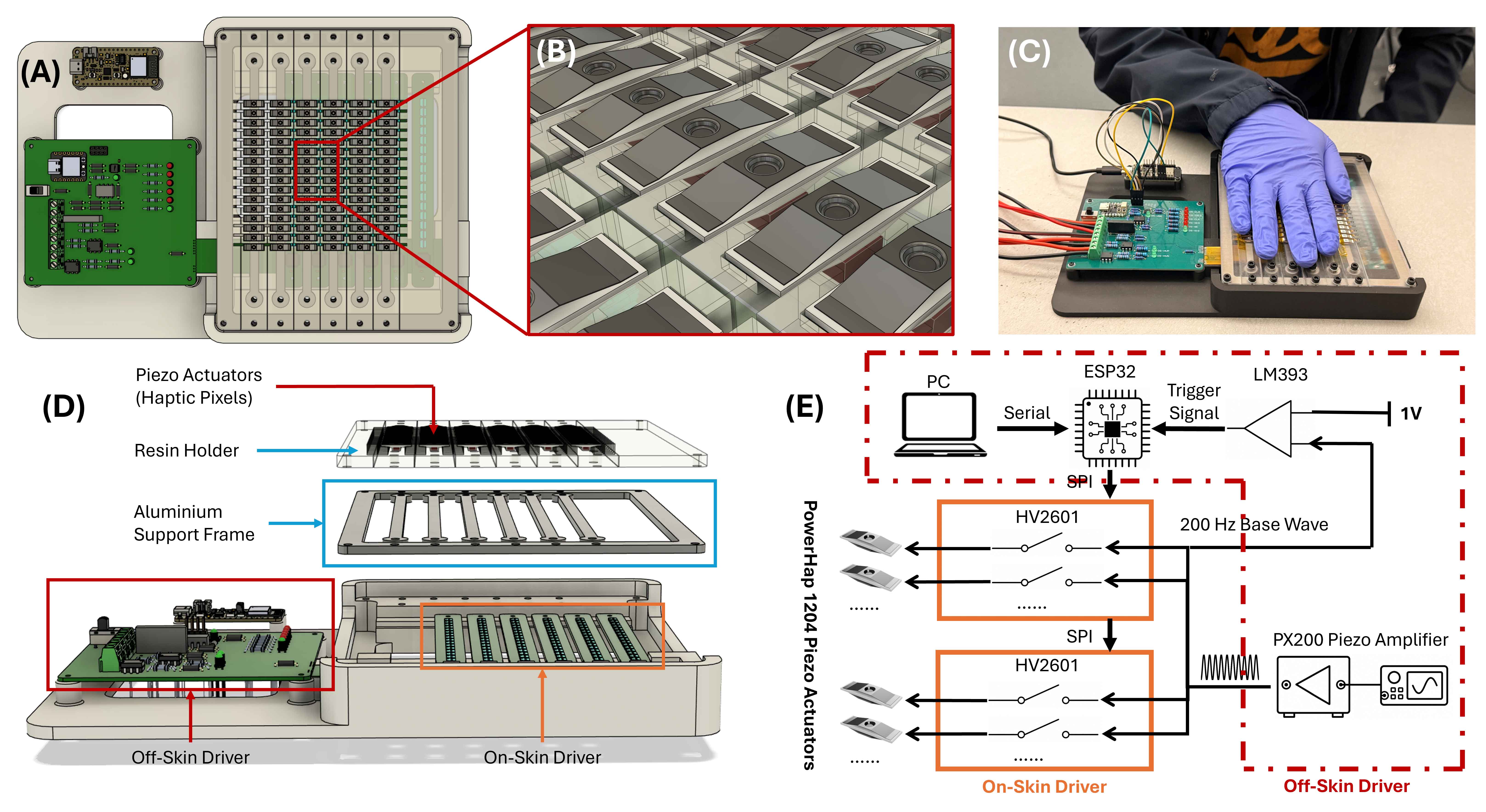}
    \caption{Implementation of the proof-of-concept prototype.
    (A) Top view of the assembled display.
    (B) Close-up of the display surface, showing the distribution of the PowerHap 1204 piezoelectric
        actuators.
    (C) Snapshot of a human-participant experiment.
    (D) Exploded view of the prototype. The piezoelectric actuators are mounted on photo-cured resin
        holders, with aluminum support frames beneath to limit deflection of the holders under hand
        pressure.
    (E) Schematic of the electrical connections. The ESP32 microcontroller sets the state of each analog
        switch channel over a high-speed SPI link, while the host PC streams haptic display frames to the
        ESP32 over a serial connection.}
    \label{fig:prototype}
\end{figure*}

Evaluating the coefficients at $m=n$ gives $a_n=-\frac{V_0}{2n}$ and $b_n=0$, so the weight of the base-frequency component decreases only as $1/n$ as the nominal frequency is lowered. A non-negligible component near $f_0$ is therefore retained even at low nominal frequencies. This property is helpful in practice, since purely harmonic vibration in the $5$ to $10\, Hz$ range typically requires displacement amplitudes on the order of hundreds of micrometers to be detected \cite{brisben1999detection}, which is difficult to achieve with small piezoelectric actuators. Preserving higher-frequency content may thus help keep low-nominal-frequency vibrations perceptible at the modest amplitudes such actuators provide.

The term at $m=1$ corresponds to the fundamental component at the nominal frequency $f_0/n$. For large $n$, its coefficients behave as
\begin{equation}
    a_1=\frac{V_0n^2\sin\frac{2\pi}{n}}{2\pi(n^2-1)}\sim\frac{V_0}{n},\quad b_1=\frac{V_0n^2\sin^2\frac{\pi}{n}}{\pi(n^2-1)}\sim\frac{\pi V_0}{n^2}
\end{equation}
so that the fundamental amplitude approaches $V_0/n$, roughly twice the magnitude of $a_n$. The component at the nominal frequency therefore remains dominant, while the residual content at the base frequency stays within the same order of magnitude. Fig.~\ref{fig:fm}C shows the computed spectra for several representative pixel intensities, in which the above trends can be observed.

We conclude this subsection with a first-order thermal analysis of SSFM, modeling each haptic pixel as a capacitive load of capacitance $C$. Two loss mechanisms are relevant. The first arises at switching events: when the voltage across the load changes abruptly by $V_j$, an energy of approximately $\frac{1}{2}CV_j^2$ is dissipated in the series resistance of the circuit. Because switching is confined to instants at which $V_b(t)<\delta$, the voltage jump satisfies $V_j\leq\delta$. For pixel intensity $n$ the switch changes state twice every $n$ base periods, so the switching rate $2f_0/n$ attains its maximum at $n=2$, that is, at the nominal frequency $f_0/2$. The second mechanism is the ohmic loss produced by the drive current flowing through the switch channel while it is closed; this contribution is bounded by that of a continuously driven pixel. Taking the worst case for both contributions, which is conservative since the two are not maximized at the same intensity, the heating power of the whole device is bounded approximately by
\begin{equation}
    H=\frac{1}{2}f_0C\delta^2N+\frac{NV_0^2R\omega^2C^2}{8(1+R^2\omega^2C^2)}
\label{eq:heat}
\end{equation}
where $N$ is the total number of haptic pixels and $R$ is the on-state resistance of each analog switch channel. Equation (\ref{eq:heat}) indicates that the switching contribution scales with $\delta^2$, so keeping the trigger threshold small keeps the overall dissipation manageable. This consideration also motivates our choice of a sinusoidal base waveform rather than a constant supply voltage $V_0$. With a constant supply, every switching event would occur at the full supply voltage, and the first term would rise to at least $\frac{1}{2}f_0CV_0^2N$. Section \ref{sec:impl} provides a numerical example in which the two cases differ by more than an order of magnitude.

\subsection{Dynamic Pattern Generation}
\label{subsec:fm-dfg}

Dynamic haptic patterns are formed by varying over time the intensity $n$ assigned to each pixel. Let the display be refreshed at a fixed interval $\Delta t$, and let the pattern rendered on a given pixel be described by an intensity sequence $\{n_p\}_{p=1}^P$. Over the interval $P\Delta t$, the drive waveform of that pixel is
\begin{equation}
    V_c(t)=\sum_{p=1}^P\mathbb{I}((p-1)\Delta t\leq t<p\Delta t)V_{c,n_p}(t)
\end{equation}
where $\mathbb{I}(\cdot)$ denotes the indicator function, which equals $1$ when its argument holds and $0$ otherwise. Assigning a different intensity sequence to each haptic pixel allows a variety of dynamic haptic textures to be rendered, such as sweeping lines, moving points, and translating shapes. Fig.~\ref{fig:fm}D illustrates how a line sweeping from left to right is composed from the intensity sequences of five pixels.

\section{Implementation Details}
\label{sec:impl}

This section describes the component selection and operating parameters of the proof-of-concept prototype. Fig.~\ref{fig:prototype}A and Fig.~\ref{fig:prototype}B show a top view and a close-up view of the assembled display. TDK PowerHap 1204 actuators were selected as the piezoelectric vibrators. Each device measures $12\times 4\, mm$ and provides a deformation amplitude of approximately $20\,\mu m$ under a $50\, V$ drive \cite{powerhap1204}. The actuators are mounted on photo-cured resin holders and bonded with 3M 467MP adhesive tape. We found cyanoacrylate adhesive unsuitable for this purpose, since variations in glue thickness introduced inconsistent damping across pixels. When a participant rests a hand on the display, the limited stiffness of the resin holders produces a visually noticeable deflection; aluminum support frames were therefore added beneath the holders to restrict this bending, as shown in Fig.~\ref{fig:prototype}D. The array comprises 16 rows and 6 columns of piezoelectric vibrators. During the experiments, participants place a hand on top of the vibrating pixels, as shown in Fig.~\ref{fig:prototype}C.

The off-skin driver is shown in Fig.~\ref{fig:prototype}E. The base waveform $V_b$ is produced by a function generator followed by a PX200 piezo amplifier, at a base frequency of $f_0=200\, Hz$. Each pixel supports seven intensity levels, $n=1,2,4,10,20,40,\infty$, corresponding to nominal frequencies of $200$, $100$, $50$, $20$, $10$, $5\, Hz$, and static; Fig.~\ref{fig:fm}C illustrates the spectral behavior using a representative set of intensities. Every $50$ ms, the host PC transmits a frame of 96 integers to an ESP32 microcontroller over a serial link, specifying the updated intensity of each of the 96 actuators. An LM393 comparator monitors the base waveform against the trigger threshold $\delta=1\, V$. When $V_b(t)<\delta$, the comparator asserts a high signal that prompts the ESP32 to begin SPI transmission. The 96 switch states must be transferred before $V_b(t)$ rises above $\delta$ again, which leaves a window of
\begin{equation}
    dt=\frac{1}{\pi f_0}\arccos\Big(1-\frac{2\delta}{V_0}\Big)=452\,\mu\mathrm{s}
\end{equation}

The SPI clock is set to $4\, MHz$, so the 96-bit transfer completes within $30\,\mu s$ and falls
comfortably inside this window.

The on-skin driver uses HV2601 16-channel high-voltage analog switches. These devices can be daisy-chained, so increasing the number of piezoelectric vibrators does not increase the number of wires running between the off-skin and on-skin drivers. With an on-state channel resistance of $R=30\,\Omega$ and a load capacitance of $C=0.5\,\mu F$, (\ref{eq:heat}) predicts a total dissipation of approximately $360\, mW$, which is manageable given the large contact area over which it is distributed. Smaller piezoelectric vibrators would present a smaller capacitance $C$, reducing the dissipation further.

For comparison, had the base waveform been a constant $50\,V$ supply, the same expression would give a total dissipation of approximately $12\,W$. Such a level would risk damaging the piezoelectric actuators and producing an unsafe surface temperature against the skin. As noted in Section~\ref{sec:fm}, this contrast is one reason a sinusoidal base waveform is preferred over a constant supply voltage.

\section{Experiments}
\label{sec:exp}

\begin{figure*}[t]
    \centering
    \includegraphics[width=\linewidth]{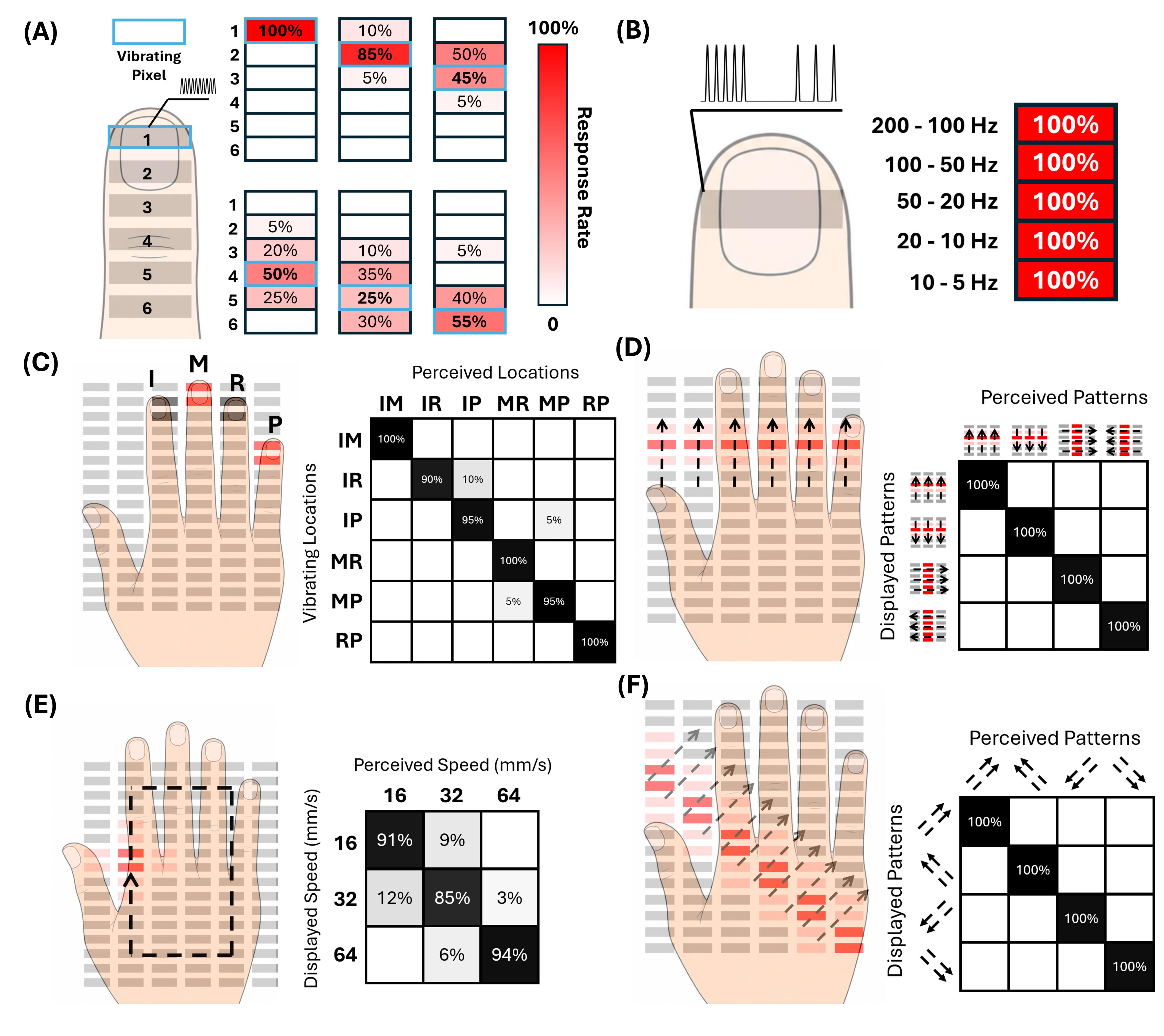}
    \caption{Human-participant experiment results. In panels C--F, rows of each matrix denote the displayed stimulus and columns the perceived response, and the entries are response rates pooled across participants. (A) Experiment 1: identification of vibration position along the index finger. Each block corresponds to one stimulated pixel, outlined in blue, and its entries give the distribution of reported positions. (B) Experiment 2: discrimination of vibration intensity. Participants reported which of two successive vibrations had the higher nominal frequency. (C) Experiment 3: multi-finger vibration perception. Participants reported which two fingers were stimulated, denoted index (I), middle (M), ring (R), and little (P). (D) Experiment 4: perception of linear motion patterns among the up, down, left, and right directions. (E) Experiment 5: perception of rotational motion speed among $16$, $32$, and $64$ mm/s. (F) Experiment 6: perception of linear motion patterns among the four diagonal directions.}
    \label{fig:exp}
\end{figure*}

We conducted six human-participant experiments to evaluate whether the prototype can convey spatial localization, frequency-encoded intensity, multi-finger stimulation, and dynamic motion patterns. Fig.~\ref{fig:exp} summarizes the experimental protocols and results. Eleven participants took part across two sessions, their ages range from 20-40 years old, with eight men and three women. Experiments 4--6 were completed by all 11 participants, while Experiments 1--3 were completed by 10 participants because one participant was unavailable for the second session. No participant received prior training.

In Experiment 1, participants placed the index finger over a row of six pixels, numbered from the fingertip (pixel 1) toward the proximal phalanx (pixel 6). In each trial, a single pixel was driven at a nominal frequency of $200$ Hz ($n=1$), and the participant reported which pixel was active. Each of the six locations was tested twice per participant, giving 20 trials per location. Fig.~\ref{fig:exp}A shows the resulting response distributions. Exact identification was highest at the fingertip, reaching $100\%$ for pixel 1 and $85\%$ for pixel 2, and declined to between $25\%$ and $55\%$ for the four more proximal pixels. The responses nevertheless remained spatially concentrated. For every stimulus location, at least $90\%$ of responses fell on the stimulated pixel or one of its immediate neighbors. Localization at the proximal sites is therefore better described as coarse than as absent. A plausible explanation for this gradient is the decrease in tactile spatial acuity from the fingertip toward more proximal regions of the finger, although the present experiment was not designed to isolate this factor.

In Experiment 2, participants attended to a single pixel that was driven sequentially at two different nominal frequencies and reported which of the two intervals had the higher frequency. Five pairs were tested, $200\,Hz$ ($n=1$) against $100\,Hz$ ($n=2$), $100$ against $50\,Hz$ ($n=4$), $50$ against $20\,Hz$ ($n=10$), $20$ against $10\,Hz$ ($n=20$), and $10$ against $5\,Hz$ ($n=40$). The order of the two intervals was randomized, and each pair was presented twice per participant, giving 20 trials per pair. As shown in Fig.~\ref{fig:exp}B, all five pairs were judged correctly in every trial. Because the tested pairs differ by a factor of two or more, this result indicates that the intensity levels available in the prototype are clearly separable. It does not establish a discrimination threshold, which would require a dedicated psychophysical study using finer frequency steps.

In Experiment 3, participants rested the index (I), middle (M), ring (R), and little (P) fingers on two pixels each. In each trial, the four pixels beneath two of the fingers were driven at $200\,Hz$, and the participant reported which pair of fingers was stimulated. Each of the six finger pairs was tested twice per participant, giving 20 trials per pair. The confusion matrix in Fig.~\ref{fig:exp}C shows correct-response rates of at least $90\%$ for all six finger pairs, reaching $100\%$ for the IM, MR, and RP pairs. In each of the few errors, one finger of the pair was reported correctly while the other was confused with an immediately adjacent finger.

Experiments 4 and 6 examined the perception of linear motion. In each trial, several rows of pixels were activated simultaneously and swept across the hand in one of four directions, selected at random. The moving front was rendered as an intensity gradient, with the pixels shown in red driven at a nominal frequency of $200\,Hz$ and those shown in pink at $50\,Hz$. Experiment 4 used the four cardinal directions (up, down, left, and right), whereas Experiment 6 used the four diagonal directions. Participants reported the perceived direction of motion. Each direction was presented twice per participant, giving 22 trials per direction. As shown in Fig.~\ref{fig:exp}D and Fig.~\ref{fig:exp}F, all trials in both experiments were answered correctly.

Experiment 5 examined the perception of motion speed. Participants were presented with a clockwise rotational pattern traversing a closed path on the palm at one of three speeds, $16$, $32$, or $64\,mm/s$, and reported the perceived speed. Each speed was presented three times per participant, giving 33 trials per speed. The confusion matrix in Fig.~\ref{fig:exp}E gives correct-response rates of $91\%$, $85\%$, and $94\%$ for the three speeds. All errors occurred between adjacent speed levels, meaning the $16\,mm/s$ pattern was never reported as $64\,mm/s$, nor the reverse. As the three levels differ by successive factors of two, this experiment likewise indicates that coarse speed differences are conveyed, without establishing the finest speed difference the display can render.

The experiments provide evidence that the proposed FM architecture preserves several forms of information that are important for a distributed haptic display despite sharing a single power-amplification source across multiple pixels. Participants could distinguish the selected frequency-encoded intensity levels, identify simultaneous stimulation across fingers, and recognize both the direction and speed of spatially propagating patterns. Importantly, the strong performance on the motion tasks (experiments 4--6) indicates that sharing the amplifier does not prevent the pixels from being coordinated with sufficient spatial and temporal independence to construct dynamic haptic patterns. The localization experiment (experiments 1--3) further suggests that the remaining errors are primarily associated with fine spatial discrimination rather than a loss of spatial organization. Even at the more difficult proximal locations, responses were concentrated on the stimulated pixel and its immediate neighbors.

These results therefore support the central feasibility claim of the SSFM approach. The proposed architecture reduces the need for pixel-level power amplification while retaining sufficient perceptual controllability to encode spatial, temporal, and intensity information in the tested conditions. This distinction is important because reducing driver hardware would provide limited benefit if the resulting shared-source architecture substantially restricted the information that could be rendered. The present experiments suggest that, at least for the coarse but practically relevant stimulus levels examined here, such a tradeoff is not inherent to the proposed design. The experiments were intentionally designed as a proof-of-concept evaluation rather than a characterization of human perceptual limits. Future psychophysical studies with finer frequency, spatial, and temporal increments will be required to determine the achievable rendering resolution and perceptual bandwidth of the architecture.

\begin{figure}
    \centering
    \includegraphics[width=0.9\linewidth]{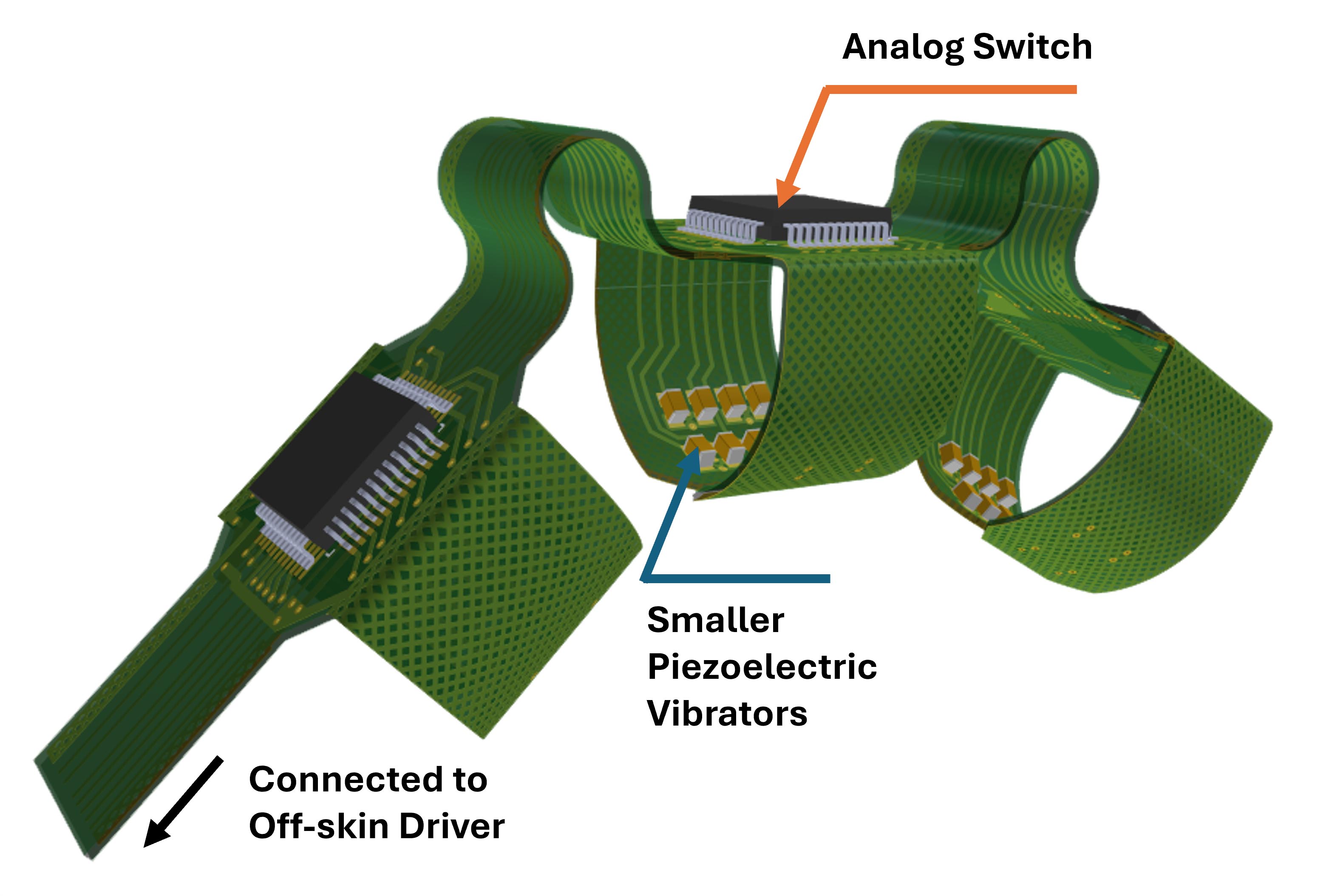}
    \caption{Rendering of a potential wearable SSFM realization on one finger. Smaller piezoelectric vibrator pieces will wrap around the user finger, and the analog switch will lie on the finger backs.}
    \label{fig:wr}
\end{figure}

\section{Discussion: Towards Wearability}
\label{sec:discussion}

Two factors govern whether the present architecture can be carried over to a dense, large-area wearable device are the footprint of the on-skin driver and the size of the piezoelectric actuators. The on-skin driver occupies a substantial area in the current prototype, as seen in Fig.~\ref{fig:prototype}D, but most of that area is taken up by the FH34D-4S connectors that interface with the PowerHap 1204 actuators rather than by active components. The HV2601 analog switches are the only components that must remain close to the pixels, and each provides 16 channels within a $7\times 7$ mm footprint \cite{hv2601}. A package of this size can be reflow-soldered onto a flexible printed circuit with a local stiffener and routed along the dorsal aspect of a finger.

On the actuator side, multilayer piezoelectric actuators with footprints as small as $2\times 2$ mm and travel on the order of $3\,\mu$m are commercially available. Actuators of this size can be soldered directly onto a flexible printed circuit, removing the need for the connectors that dominate the present board. Although this travel is considerably smaller than the $20\,\mu$m provided by the PowerHap 1204 actuator, the FM encoding described in Section~\ref{sec:fm} is intended precisely for this regime, since perceptual intensity is carried by vibration frequency rather than by displacement amplitude.

These observations outline a possible engineering path. An on-skin layer formed by a flexible PCB carrying both the HV2601 switches and directly mounted millimeter-scale actuators, connected to the off-skin amplifier by the small, fixed number of conductors needed for the shared base waveform and the daisy-chained control bus. At these dimensions, the switching circuitry would occupy roughly $3$ mm$^2$ per pixel, so the driver need not dominate the area budget of the on-skin layer. Equation (\ref{eq:heat}) further suggests that the smaller capacitance of miniature actuators would reduce the heat dissipation per pixel, partly offsetting the higher pixel count. Fig. \ref{fig:wr} gives a rendering of the possible engineering realization on one finger.

We stress that these are design projections rather than demonstrated results. A practical device would still need to address mechanical coupling to the skin and the perceptual consequences of a reduced displacement amplitude. We leave such an implementation to future work.

\section{Conclusion}
\label{sec:conclusion}

We investigated a frequency-modulated approach to distributed haptic display, in which perceived intensity is encoded through vibration frequency rather than displacement amplitude, and developed the SSFM structure, where a single high-voltage amplifier supplies a common base waveform to all pixels while compact analog switches set the nominal frequency of each pixel independently. The amplifier thereby belongs to the off-skin driver and the on-skin driver reduces to switching circuitry and interconnections, addressing the per-pixel amplification and routing that grows as amplitude-modulated architectures scale. Spectral and thermal analyses of the gated waveform guided the choice of a sinusoidal base waveform and a low switching threshold. In six experiments with untrained volunteers on a 96-pixel prototype, responses to single-pixel stimulation fell on the stimulated pixel or an immediate neighbor in at least $90\%$ of trials, nominal frequency pairs and linear sweep directions were reported without error, finger pairs were identified with at least $90\%$ accuracy, and rotational speed with $85\%$ to $94\%$ accuracy, with all errors confined to adjacent levels. These results support the feasibility of FM encoding under a shared amplifier although the stimulus levels were coarse. Measuring discrimination vibration amplitude thresholds and realizing the wearable implementation outlined in Section~\ref{sec:discussion} are our next steps.

\section*{Acknowledgements}

\textbf{AI Usage}: ChatGPT was used to assist with language polishing. Claude Code was used to assist with firmware coding. The research design, hardware development, experiments, data analysis, figures, and scientific conclusions were completed by the human authors.

\addtolength{\textheight}{-12cm}   

\bibliographystyle{IEEEtran}
\bibliography{references}

\end{document}